\documentclass[runningheads]{llncs}

\usepackage[review,year=2024,ID=7398]{eccv}
\usepackage{eccvabbrv}
\usepackage{graphicx}
\usepackage{booktabs}
\usepackage[accsupp]{axessibility}  
\usepackage[pagebackref,breaklinks,colorlinks,citecolor=eccvblue]{hyperref}
\usepackage{amsmath}
\usepackage{graphicx}
\usepackage{multirow}

\usepackage{amsmath,amsfonts,bm}

\def\eqref#1{equation~\ref{#1}}

\def\1{\bm{1}}

\DeclareMathAlphabet{\mathsfit}{\encodingdefault}{\sfdefault}{m}{sl}
\SetMathAlphabet{\mathsfit}{bold}{\encodingdefault}{\sfdefault}{bx}{n}

\def\gD{{\mathcal{D}}}

\usepackage{orcidlink}
\usepackage{xcolor}
\usepackage{longtable}
\usepackage{supertabular}
\newcommand{\CUT}[1]{}

\newcommand{\tabincell}[2]{\begin{tabular}{@{}#1@{}}#2\end{tabular}} 

\begin{document}

\title{Borrowing Treasures from Neighbors: In-Context Learning for Multimodal Learning with Missing Modalities and Data Scarcity } 
\author{Zhuo Zhi\inst{1}\and
Ziquan Liu\inst{2}  \and
Moe Elbadawi \inst{3} \and
Adam Daneshmend \inst{4} \and
Mine Orlu \inst{5} \and
Abdul Basit \inst{5} \and
Andreas Demosthenous \inst{1}\and
Miguel Rodrigues \inst{1}
}

\institute{Department of Electronic and Electrical Engineering, UCL, UK \and
School of Electronic Engineering and Computer Science, QMUL, UK \and
School of Biological and Behavioural Sciences, QMUL, UK \and
University College London Hospitals NHS Foundation Trust, UK\and
UCL School of Pharmacy, UCL, UK
}




\maketitle

\begin{abstract}
Multimodal machine learning with missing modalities is an increasingly relevant challenge arising in various applications such as healthcare. This paper extends the current research into missing modalities to the low-data regime, i.e., a downstream task has both missing modalities and limited sample size issues. This problem setting is particularly challenging and also practical as it is often expensive to get full-modality data and sufficient annotated training samples. We propose to use retrieval-augmented in-context learning to address these two crucial issues by unleashing the potential of a transformer's in-context learning ability. Diverging from existing methods, which primarily belong to the parametric paradigm and often require sufficient training samples, our work exploits the value of the available full-modality data, offering a novel perspective on resolving the challenge. The proposed data-dependent framework exhibits a higher degree of sample efficiency and is empirically demonstrated to enhance the classification model's performance on both full- and missing-modality data in the low-data regime across various multimodal learning tasks. When only 1$\%$ of the training data are available, our proposed method demonstrates an average improvement of 6.1$\%$ over a recent strong baseline across various datasets and missing states. Notably, our method also reduces the performance gap between full-modality and missing-modality data compared with the baseline.  Code is available\footnote{\href{https://github.com/ZhuoZHI-UCL/ICL_multimodal}{GitHub repository}}.


\end{abstract}
\section{Introduction}
\vspace{-0.3cm}
\label{sec:intro}

\begin{figure*}[t]
    \centering
    \includegraphics[width=1\linewidth]{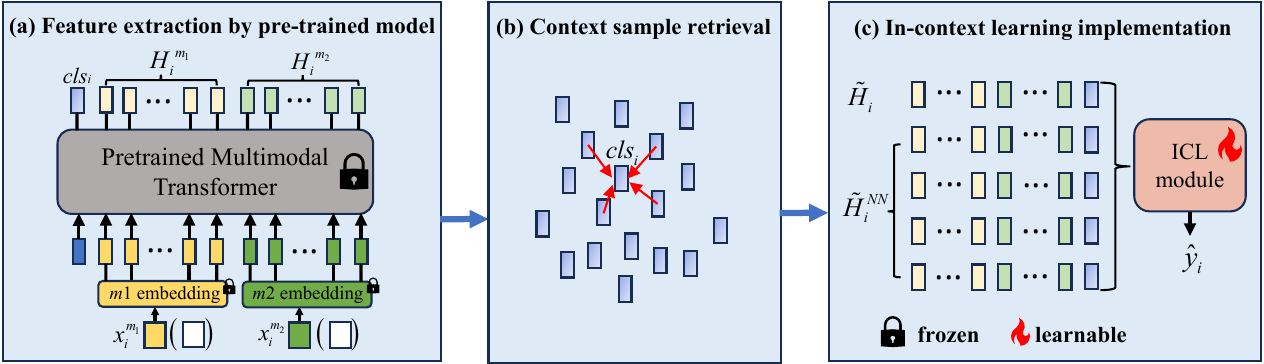}
    \vspace{-0.7cm}
    \caption{The overview of the proposed method. (\textbf{a}) Assuming that each sample contains data with 2 modalities $x_i^{m_1}$ and $x_i^{m_2}$, we get the feature $H_i = ({H_i}^{m_1},{H_i}^{m_2},cls_i)$ of the sample by using a pre-trained multimodal transformer, note that $x_i^{m_1}$ or $x_i^{m_2}$ may be missed. (\textbf{b}) We use the $cls$ token to calculate the cosine similarity between the current sample and all full-modality training samples, and then retrieve the most similar $Q$ samples. (\textbf{c}) We input the pooled feature of the current sample  $\tilde{H}_i$ and neighbor samples $\tilde{H}^{NN}_i$ into the ICL module to predict the label ${\hat{y}}_i$. Note that only the ICL module requires to be trained and the others are frozen. The retrieval-augmented operation is the same for both the training and inference processes. Note that the words ‘missing modality' and 'incomplete modality', 'full modality' and 'complete modality' are used interchangeably.}
    \vspace{-0.6cm}
    \label{fig:overall_framework}
\end{figure*}

Humankind leverages multimodal data to make intelligent decisions, such as vision, language and sound \cite{baltruvsaitis2018multimodal}. Consequently, multimodal machine learning (ML) has emerged as a pivotal learning paradigm in the ML research community, aiming to improve the quality of decision-making by using multimodal data in various fields, e.g., ML-assisted healthcare \cite{hayat2022medfuse} and malicious content detection \cite{kiela2020hateful}. However, a major challenge in the application of multimodal ML is the missing-modality issue \cite{suo2019metric,tsai2018learning,suo2019metric,ma2022multimodal}, where some data samples do not have complete modalities due to challenges in the data collection process. For instance, in medical applications, some modalities, such as X-ray images \cite{johnson2019mimic}, are more expensive and/or time-consuming to obtain than others, e.g., Electronic Health Records (EHRs).  \cite{johnson2023mimic}. Therefore, a multimodality dataset with the missing-modality issue contains samples with complete modalities, i.e., \emph{full-modality data}, and also samples with incomplete modalities, i.e., \emph{missing-modality data}. 

\begin{figure}[t]
    \centering
     \includegraphics[width=1.0\linewidth]{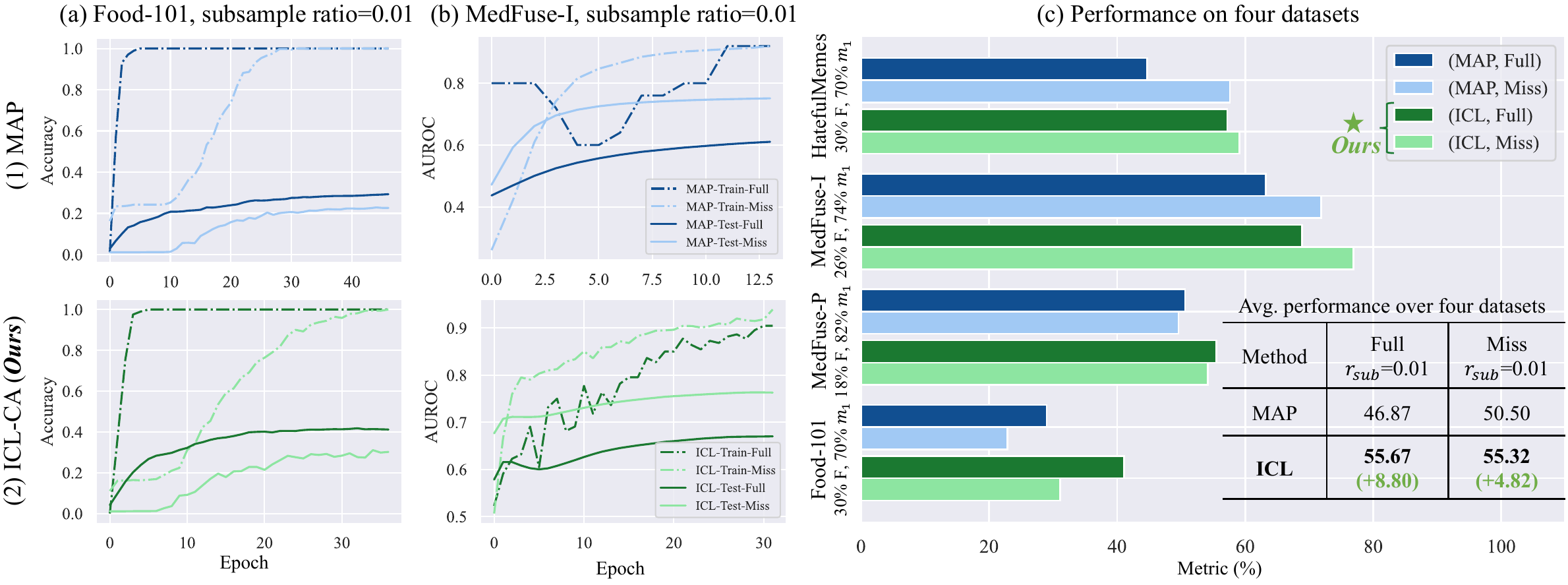}
     \vspace{-0.7cm}
    \caption{(a) The learning curve of ICL-CA (ours) and Missing-Aware Prompt (MAP) \cite{lee2023multimodal} on the Food-101 dataset in the low data regime. (b) The learning curve of two methods on the MedFuse-I dataset. The subsampling ratio is set to be 0.01. The difference in the learning steps is due to early stopping.
    During each training process, we calculate the metric of missing- and full-modality samples \emph{separately} and refer to them as '-full' and '-miss'. (c) The performance of MAP and our ICL-CA on four multimodality datasets with the missing-modality issue. The y-axis shows the dataset name and missing status. The x-axis is metrics for each dataset, AUROC for MedFuse-I, MedFuse-P and HatefulMemes, and accuracy for Food-101.
    On each dataset, we compute the metric for test data with full and missing modalities \emph{separately} and show the results in \textbf{dark} and \textbf{\textcolor{gray}{light}} color. The legend means \emph{(Method, Full/Missed-Modality)}. When the task complexity is low, e.g., binary classification tasks like HatefulMemes \cite{kiela2020hateful} and MedFuse-I \cite{hayat2022medfuse}, the performance of full-modality data lags behind that of missing-modality data, as fitting the training data does not need full-modality information. When the task complexity is high, e.g., a multi-classification task like Food-101 (101 classes) \cite{bossard14}, the full-modality performance surpasses that of the missing-modality, as the task requires all modalities to adequately model the training data. Our ICL is significantly better than MAP on four datasets in four cases as the table below shows. See more details in Sec.~\ref{sec:Experiment}.}
    \label{fig:intro}
    \vspace{-0.7cm}
\end{figure}

Existing research on tackling the missing-modality challenge has two pathways. Before the advent of multimodal transformer \cite{vaswani2017attention,kim2021vilt}, multimodal learning relies on explicit information fusion with features, the output of modality-dependent backbones. Thus, some research work \cite{ma2021smil} proposes to learn a parametric model to infer missing modalities. In the era of multimodal transformers, the modality fusion starts from the input layer as a single transformer can handle various input formats, such as vision, language and sound \cite{radford2023robust}. A recent work \cite{lee2023multimodal} proposes to learn the maximum likelihood estimation for missing modalities at the token level. However, there are two limitations in existing research. Firstly, it is frequently assumed that the sample size during training is adequate so a parametric model can be learned to estimate the missing modalities \cite{ma2021smil,lee2023multimodal}, but the sample size is not always sufficient in the real-world \cite{huang2021makes}. Secondly, there is a notable absence of analysis concerning the performance disparity between missing- and full-modality data on multimodal learning.

To further our understanding of the missing-modality challenge in the low-data regime, our paper analyzes the performance of missing- and full-modality data separately in various tasks and training sample sizes. There are two major observations and hypotheses in this paper: 1) The performance of existing methods drops significantly in the low-data regime, and the potential of limited data needs to be more fully exploited. 2) Different tasks depend on full- and missing modality data to different degrees. For low-complexity tasks, the model mainly learns from missing-modality data \cite{vale2021long,wang2020makes}, leading to higher performance on missing-modality data compared with full-modality ones. In contrast, the model performs worse on missing-modality data compared with full-modality ones in high-complexity tasks. Therefore, we should not only focus on reconstructing/improving missing-modality data. See Fig.~\ref{fig:intro} for the empirical evidence. 

Motivated by our empirical observation, we propose a data-dependent approach based on \emph{retrieval-augmented in-context learning} (ICL) \cite{borgeaud2022improving,ram2023context}, to reduce the performance drop of multimodal learning with missing modality in the low-data regime.  The proposed method exploits the value of available data and adaptively enhances both missing- and full-modality samples by using the neighboring full-modality samples, as Fig.~\ref{fig:overall_framework} shows. Specifically, we train an ICL module on top of the features of a frozen multimodal transformer, such as ViLT \cite{kim2021vilt}, whose context is full-modality data retrieved from the training set using the cosine-based similarity measure. In cases where modalities are missing, their features are integrated with similar full-modality features, enabling the model to implicitly infer the absent modality for enhanced performance on the target task. For full-modality data, feature refinement is performed using neighbor information to optimize prediction accuracy. Consequently, the ICL module demonstrates improved performance on both missing- and full-modality data across hard and easy tasks, while concurrently diminishing the performance disparity between the two data types. Our experiments validate the effectiveness of the proposed ICL method on various datasets with extensive experiments, see Fig.~\ref{fig:intro} and Sec.~\ref{Main Results}. Our main contributions are three-fold:
\begin{enumerate}
    \item We investigate the data scarcity issue in missing-modality tasks and unveil the drawback of the existing parametric approach in the low-data regime, as its effectiveness often relies on a sufficient sample size. Our empirical study also reveals that the model should adaptively focus on two types of data
    as dependence on missing-modality data is not necessarily worse than that of full-modality ones.
    \item We propose a novel data-dependent in-context learning method to improve the sample efficiency and benefit the learning of both missing- and full-modality data,  where the nearest neighbor information of full-modality data is exploited. To the best of our knowledge, our work is
among the first to use in-context learning to address
the challenge of missing modality in the low-data regime. 
    \item Our experiment demonstrates the effectiveness of the proposed ICL method on four datasets, including both medical and vision-language multimodal learning tasks. The averaged performance gain on four datasets over the baseline MAP in the low-data regime is 6.1$\%$.
\end{enumerate}
 The paper is organized as follows. Sec.~\ref{sec:related} gives an overview of related research. Sec.~\ref{sec:method} first introduces our empirical observation about the performance gap and learning process difference between missing- and full-modality data, and then elaborates on the proposed method. Sec.~\ref{sec:Experiment} shows the experiment results and the analysis. Finally, Sec.~\ref{sec:conclusion} summarizes this paper, its limitations and future directions.
\vspace{-0.3cm}

\section{Related Work}
\label{sec:related}
\vspace{-0.3cm}

\noindent\textbf{Missing Modalities in Multimodal Learning.}
Multimodal models usually assume that the input samples have complete modalities. However, the problem of missing modalities exists in various applications. In ML for healthcare, combining EHR and X-ray as input excels in mortality prediction and phenotype classification task \cite{hayat2022medfuse}, but some patients do not have the time or financial support for X-rays. In ML with vision-language data, a model may not receive image input from the user as a result of network/format issues \cite{lee2023multimodal}. 
The model fails to perform as expected in these situations \cite{ma2022multimodal}.
Consequently, much work has been devoted to improving the robustness of multimodal models under modal absence. \cite{tsai2018learning} optimizes a joint generative-discriminative objective for multimodal data and labels which contain the information required for generating data for missed modality. In \cite{ma2021smil}, multimodal learning with severely missing modalities (SMIL) is designed to reconstruct the missing modalities using modality priors and Bayesian
Meta-Learning.
\cite{lee2023multimodal} introduced two types of missing-aware prompts that can be seamlessly integrated into multimodal transformers. 
\cite{ma2022multimodal} proposes a unified strategy based on multi-task optimization to deal with missing modalities in the transformer-based multimodal model. \cite{zeng2022tag} proposes a tag encoding module to assist the transformer’s encoder learning with different missing modalities. In much of the existing work, the parametric approach is adopted, which learns a model to handle samples with missing modalities and only uses that model to infer the missed modalities during the test stage. Its drawback is that learning such a model requires sufficient training data so the parametric approach cannot perform well when the sample size is low, see Sec.~\ref{Main Results}. In contrast, our paper uses the semi-parametric approach to deal with the missing-modality issues. In our approach, we enhance the current data by retrieving similar samples during both the training and inference phases. This strategy emphasizes the quality of the samples, thereby reducing reliance on sample size.

\noindent\textbf{Data scarcity in transfer Learning.}
Multimodal learning based on pre-training and fine-tuning has become popular \cite{xu2023multimodal,kim2021vilt,radford2021learning,li2021align,wang2022image}. The performance of the pre-trained model on the target task is highly dependent on the size of the fine-tuning dataset, which is particularly problematic in certain scenarios.  For example, scarce positive samples are recorded for some rare diseases \cite{mazurowski2008training}, or, a small amount of text data is available for some low-resource language tasks \cite{hedderich2021survey}. Many existing works focus on facilitating transfer learning under low-data situations. \cite{evci2022head2toe} proposes to select features from
all layers of the source model to train a classification head for the target domain, which matches performance obtained with fine-tuning on average while reducing training and storage costs. Similarly, \cite{zhang2022fine} introduces the algorithm for the large-scale pre-trained models during low-data fine-tuning, which adaptively selects a more promising subnetwork to perform staging updates based on gradients of back-propagation. From the data perspective, \cite{liu2022improved} proposes a novel selection strategy to select a subset from pre-training data to help improve the generalization on the target task. Likewise, the prototypical fine-tuning approach is proposed in \cite{jin2023prototypical}, which automatically learns an inductive bias  to improve predictive performance for varying data sizes, especially low-resource settings. In contrast, we explore the application of a data-centric approach within the domain of multimodal learning, specifically addressing scenarios involving missing modalities and data scarcity. Our work demonstrates the effectiveness of the data-centric approach in this novel domain.
\CUT{
\begin{figure}[t]
    \centering
    \includegraphics[width=1\linewidth]{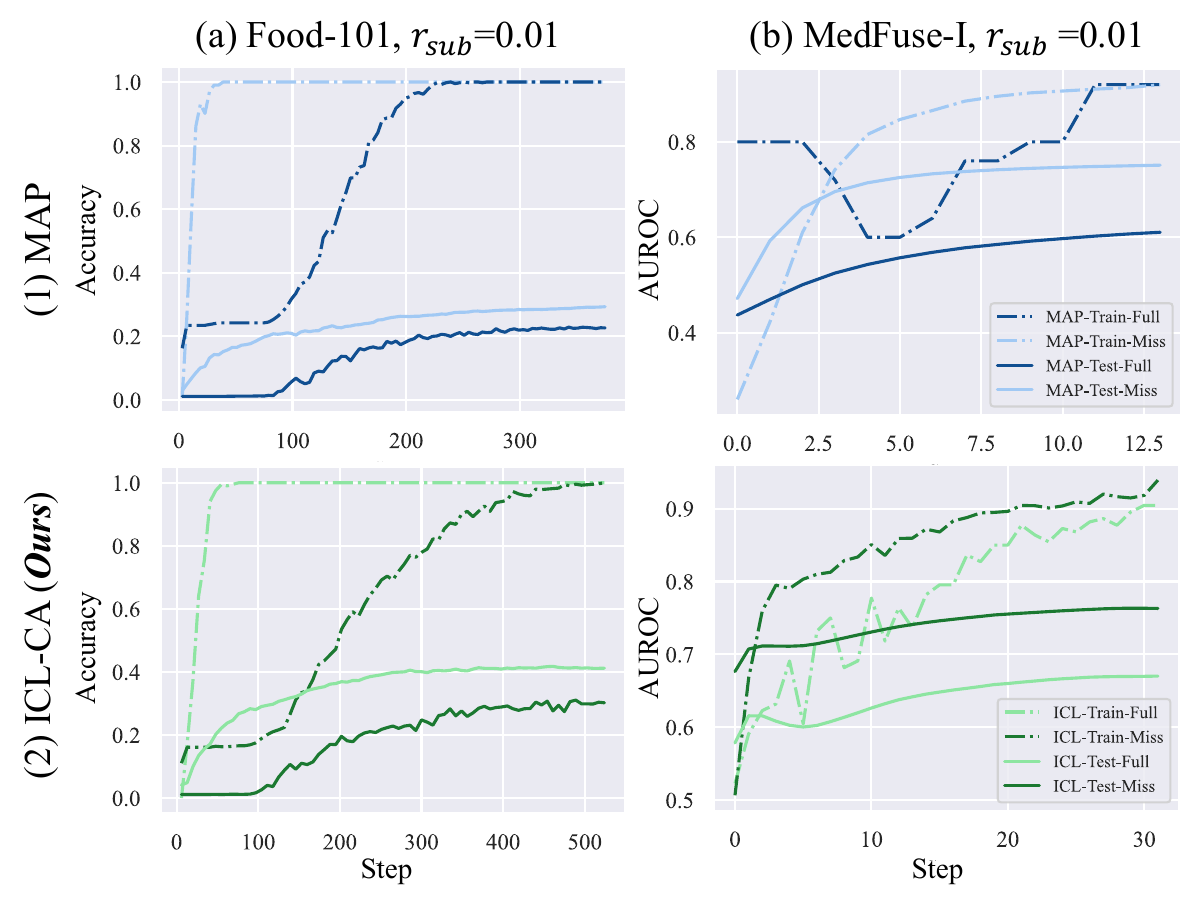}
    \vspace{-0.8cm}
    \caption{The learning curve of ICL-CA (cross attention) and MAP on different datasets under low data regime. (a) The learning curve of two methods on the Food-101 dataset. (b) The learning curve of two methods on the MedFuse-I dataset. The subsampling ratio is set to be 0.01. The difference in the learning steps is due to early stopping, indicating that MAP tends to overfit earlier than ICL-CA.}
    \vspace{-0.6cm}
    \label{fig:learning_curve}
\end{figure}
}

\noindent\textbf{In-context learning (ICL).}
ICL has emerged as a potent transfer learning approach in natural language processing (NLP), where large language models (LLMs) leverage context augmented with a few examples to make predictions, circumventing the need for parameter updates typical in supervised learning \cite{dong2022survey}. Demonstrating versatility, LLMs apply ICL to perform complex tasks, including mathematical reasoning and commonsense answering \cite{wei2022chain}. The success of ICL in NLP has recently spurred its adoption in diverse modalities, such as visual \cite{bar2022visual,wang2023images, wang2023seggpt, gupta2023visual}, speech \cite{wang2023neural,zhang2023speak}, and multimodal domains \cite{alayrac2022flamingo, huang2023language, hao2022language, koh2023grounding, tsimpoukelli2021multimodal}. In the work of \cite{tsimpoukelli2021multimodal}, a vision encoder trained on aligned image-caption data represents images as sequences of continuous embeddings. This approach, using a frozen language model, surprisingly adapts to new tasks through ICL conditioning. Similarly, Flamingo \cite{alayrac2022flamingo}, trained on extensive multimodal web corpora, showcases few-shot learning capabilities via ICL. Our paper specifically targets scenarios characterized by missing modalities and limited data, aiming to harness contextual features from full-modality samples. To the best of our knowledge, our work is
among the first to use in-context learning to address such a challenge, offering a novel perspective on improving sample efficiency and reducing the performance gap between missing- and full-modality.



\vspace{-0.3cm}
\section{Proposed Method}
\label{sec:method}
\vspace{-0.2cm}
We first describe the problem definition and then the existing baseline to handle the missing-modality issue. Our empirical observation and the proposed method are elaborated on later. 
\vspace{-0.4cm}
\subsection{Problem Setting}
\vspace{-0.3cm}
We consider the multimodal learning problem with a dataset $\gD$ containing multimodal input samples, $\gD$ can be the training/validation/testing dataset. For notation simplicity, we assume there are two modalities in the dataset, i.e., $\gD=\{x_{i}^{m_{1}}, x_{i}^{m_{2}}, y_{i}\}_{i=1}^{N}$ where the label $y_{i}\in\{1,\cdots, K\}$, but note that our framework can handle any number of modalities in principal. It is assumed that some samples have missing data for a particular modality. For example, some patients do not have the time or financial support for X-rays, and some images from food reviews failed to upload due to network/format issues. More importantly, we assume that the training set size $N$ is limited as a result of the complicated data collection process \cite{johnson2023mimic} and expensive human-expert annotations \cite{medmnistv2}. The prevalence of missing modalities and limited data within our problem context is a common occurrence in critical domains, such as medical data analysis, thus requiring immediate resolution \cite{wang2023prototype,zhang2022m3care}.



\noindent\textbf{Parametric Approach to Mitigating Missing Modalities}. 
As described in Section \ref{sec:related}, parametric methods are employed widely for handling missing modalities in multimodal learning. We introduce this type of approach using a representative work--missing aware prompt (MAP)\cite{lee2023multimodal}.  In this method, empty prompt tokens are initialized and concatenated with the input sequences for each layer of the multimodal transformer. In this way, the prompts for the current layer can interact with the prompt tokens inheriting from previous layers, and thus learn more effective instructions for the model prediction. Only the parameters of the prompts and the final classification layer are updated during training.


\subsection{Empirical Observations in the Low-Data Regime}
Fig.~\ref{fig:intro}-a1 and \ref{fig:intro}-b1 show the learning curve of MAP on two datasets with missing modalities. A discernible divergence is evident in the training curve of full versus missing modalities in the two tasks. For a relatively straightforward task, i.e., MedFuse-I (binary classification task), the training AUROC of full-modality data is lower than that of missing-modality data in many learning steps. In contrast, for the more complex Food-101 dataset (a multi-classification task with 101 classes), the trend is reversed, where the full-modality data have better performance. This observation implies that whether missing-modality data are harder to learn than full-modality data depends on the task complexity. Note that although a similar observation is shown in \cite{wang2020makes}, our work contributes by unveiling this phenomenon in the training of a multimodal transformer model, instead of the joint encoder training in \cite{wang2020makes}.
This insight leads us to conjecture that \textit{only focusing on reconstructing information for the missing modalities is not an optimal solution}, as the missing-modality data are not necessarily more difficult to learn than full-modality data. Consequently, we propose an ICL-based approach, where each sample, regardless of its modality completeness, \textbf{adaptively} benefits from its fusion with neighbor full-modality samples. The benefit of ICL is demonstrated in the ICL learning curve of Fig.~\ref{fig:intro}-a2 and \ref{fig:intro}-b2, where the generalization of both data types is improved. 
\vspace{-0.3cm}

\subsection{Borrowing Treasures from Your Neighbors: A Semi-Parametric Approach}

Unlike parametric methods, we adaptively augment full- and missing-modality samples through in-context learning by a limited number of parameters, which fully exploits the value of available data.

\noindent{\textbf{The proposed Borrowing Treasures from Your Neighbors method.}} In-context learning enables LLMs to perform tasks by conditioning an input prompt with exemplar examples without the need for parameter optimization. Drawing inspiration from it, we introduce the method titled \emph{Borrowing Treasures from Your Neighbors}. This approach leverages similar data with full modalities to improve the performance on data containing full and missing modalities, aiming to alleviate the challenges posed by missing modalities and data scarcity. The reason why we only retrieve full-modality training data is that the missing-modality data need a reference of full modalities to implicitly infer the missed modalities, and the full-modality features can be fused with other full-modality ones to improve the generalization. Tab.~\ref{tab: retrieving groups exp} shows the strength of only using full-modality neighbors compared with using all training data and only missing-modality data.

 As shown in Fig. \ref{fig:overall_framework}, for each sample $\left\{x_{i}^{m_{1}}, x_{i}^{m_{2}}, y_{i}\right\}$, the pre-trained multimodal transformer $f_{\theta_{\text{X}}}$ infers the feature $H_i$ including features for $x_{i}^{m_{1}}$, $x_{i}^{m_{2}}$ and a CLS feature $cls_i$ by $H_i = \{{H_i}^{m_1},{H_i}^{m_2},cls_i\} = f_{\theta_\text{X}}(x_{i}^{m_{1}}, x_{i}^{m_{2}}), {H_i}^{m_1} \in \mathbb{R}^{L_1 \times d}, {H_i}^{m_2} \in \mathbb{R}^{L_2 \times d}, cls_i\in \mathbb{R}^{1\times d}$, where $L_1$ and $L_2$ are the number of tokens of embedded $x_{i}^{m_{1}}$ and $x_{i}^{m_{2}}$, respectively. $d$ is the embedding dimension.
 Based on the extracted features, the most common approach is to directly train a classifier $f_{{\theta}_{\text{c}}}$ to predict the labels $\hat y_i = f_{{\theta}_{\text{c}}}({H_i}^{m_1},{H_i}^{m_2},cls_i)$. 
When there is missing-modality data, we follow MAP \cite{lee2023multimodal} to use default tokens to fill those missing tokens, see Sec.~\ref{sec:Experiment} for details. Then we retrieve the most similar $Q$ samples ${H_i}^{NN} = \{H_{i,q}^{m_1},H_{i,q}^{m_2},cls_{i,q}\}_{q=1}^{Q}$ from the training samples with full modalities, where the features are arranged in descending order according to the similarity. The similarity is determined by cosine similarity and calculated with the $cls$ tokens in our implementation.
Finally, we design the ICL module to predict labels from the mean-pooled feature $
\tilde H_i$ of the current sample and the mean-pooled retrieved \emph{context} $\tilde{H}^{NN}_i$.



\subsection{In-context module design}\label{In-context module design}
Our proposed method does not update/add any parameters in the pre-train multimodal model. During the training phase, we freeze all the parameters $f_{\theta_{\text{X}}}$ of the multimodal transformer (including the input embedding layers).  We only update the parameters of the ICL module.

Next, we introduce the details of the ICL module. Transformer-based structures are shown to be capable of in-context learning \cite{brown2020language, dong2022survey}. Inspired by them, we compare two configurations of ICL based on transformer:   ICL by cross-attention and ICL by next-token prediction. For ease of explanation,  we assume that $N =2$ and use blocks with different colors to represent tokens for different modalities.

\begin{figure} [t]
        \centering
        \vspace{-0.2cm}
	\includegraphics[width=0.8\linewidth]{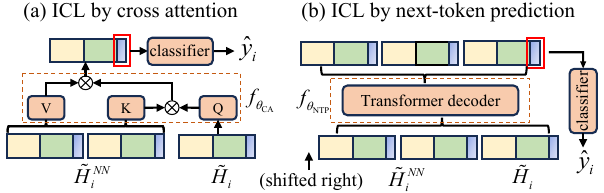}
 \vspace{-0.2cm}
	\caption{The illustration of two ICL approaches.  (a) ICL by cross attention.
 (b) ICL by next-token prediction. The yellow and green tokens denote features of two different modalities and the blue token is the $cls$ token.} 
 \vspace{-0.5cm}
	\label{design_of_ICL}
\end{figure}

\noindent{\textbf{ICL by cross attention (ICL-CA).}} One approach to perform ICL is to update the current sample's feature using the cross attention with nearest neighbor (NN) samples as keys, as Fig.~\ref{design_of_ICL}(a) shows. The cross attention function $f_{\theta_{\text{CA}}}$ is trained to minimize the classification loss using the classification token, i.e.,  
\begin{align}
    \hat{cls}_i&=f_{\theta_{\text{CA}}}(\tilde{H}_i,\tilde{H}^{NN}_i), \ell_{\text{CA}}^{(i)}=\ell_{\text{cls}}(\hat{cls}_i,y_i),
\end{align}

\CUT{
\begin{equation}
Q=\tilde{H}_i W^Q;
K=\tilde{H}_i^{N N} W^K;
V=\tilde{H}_i^{N N} W^V,
\end{equation}
where $W^Q, W^K$ and $W^V$ are weight matrices to be learned. Then, we define the function for the ICL module as
\begin{align}
&\hat{y}_i = clf_{\theta_{ICL}}(O_i),\\
&O_i = \operatorname{Softmax}\left(\frac{Q K^T}{\sqrt{d_k}}\right)V,
\end{align}
}

where $\ell_{\text{cls}}$ means a classification loss such as cross-entropy and $\ell_{\text{CA}}^{(i)}$ is the loss value for the $i$th sample by ICL-CA method. In the cross attention module, the sample interacts with the tokens from similar full-modality samples, and thus implicitly infers missing modalities for missing-modality samples or refines the features for full-modality ones. We give more details of ICL-CA in Section \ref{ICL module setting.}.  

\noindent{\textbf{ICL by next-token prediction (ICL-NTP).}}
Another way to apply ICL is to implement the next-token prediction by transformer decoder, which is shown in Fig.~\ref{design_of_ICL}(b). Write the input of the transformer decoder $[\tilde{H}_i^{N N};\tilde{H}_i]$ as

\begin{align}
[h_{i,1}^{(1)},\cdots,h_{i,1}^{(T)},cls_{i,1};&\cdots;h_{i,Q}^{(1)},\cdots,h_{i,Q}^{(T)},cls_{i,Q};\\\nonumber
&h_{i,Q+1}^{(1)},\cdots,h_{i,Q+1}^{(T)},cls_{i,Q+1}],\nonumber
\end{align}

where $h_{i,q}^{(t)}, q \in \{1, \ldots, Q+1\}, t \in \{1, \ldots, T\} $ is the $t$th token of the $q$th neighbor (the $Q+1$th neighbour is the current sample itself). $cls_{i,q}, q \in \{1, \ldots, Q+1\}$ is the $cls$ token of the $q$th neighbor.   The decoder function $f_{\theta^{\text{(NTP)}}}$ is trained to predict the next token in an auto-regressive way, i.e.,
\begin{align}
    &\hat h_{i,q}^{(t)}/\hat{cls}_{i,q}^{(t)} = f_{\theta_{\text{NTP}}}(h_{i,1}^{(1)},\cdots, h_{i,q}^{(t-1)}),\\
    &\ell_{\text{NTP}}^{(i)}=\lambda_{\text{NTP}}\sum_{q=1}^{Q+1}\sum_{t=1}^T(\hat h_{i,q}^{(t)}-h_{i,q}^{(t)})^2+\sum_{q=1}^{Q+1}\ell_{\text{cls}}(\hat{cls}_{i,q},y_{i,q}),
\end{align}
where $\lambda_{\text{NTP}}=0.1$ is an adjustable hyperparameter to introduce the loss of feature reconstruction. $y_{i,q}$ is the label of the $q$th neighbour. $\ell_{\text{NTP}}^{(i)}$ is the loss value for the $i$th sample by ICL-NTP method. While $\hat{cls}_{i, Q+1}$ is used to predict $\hat{y}_i$, we incorporate other tokens ($h$'s) in the loss computation to improve the prediction ability of ICL-NTP. This approach ensures that the outcomes of prior predictions continuously inform the subsequent token prediction, compelling the current sample to assimilate the rich context provided by its neighbors, i.e., each prediction is learned from the accumulation of preceding ones. The detailed settings are described in Section~\ref{ICL module setting.}. Note that the reason why we explore different ICL configurations is to provide a comprehensive understanding of the in-context learning in our problem setting.


\vspace{-0.5cm}

\section{Experiment}
\label{sec:Experiment}

We first introduce the experimental settings and then present the experimental results of our methods and baselines on four datasets, demonstrating the effectiveness of our method in missing modality and low-data tasks.

\subsection{Experimental Setting}
\noindent{\textbf{Datasets.}} We follow existing works using two-modality datasets for a standard comparison \cite{lee2023multimodal}. Specifically,
we use two medical multimodal datasets containing EHR and X-ray images, i.e., MedFuse-In-hospital mortality (MedFuse-I) \cite{hayat2022medfuse} and MedFuse-Phenotype (MedFuse-P)  \cite{hayat2022medfuse}, and two general vision-language datasets (UPMC Food-101 \cite{bossard14} and Hateful Memes \cite{kiela2020hateful}) in our experiment. The details of each dataset are in Appendix A. 

\noindent{\textbf{Baseline.}} We compare our method with the most recent baseline to tackle the missing-modality issue in multimodal transformers, i.e., missing-aware prompts (MAP) \cite{lee2023multimodal}. We also compare with two methods commonly used in transfer learning: 1) fine-tuning all layers of the pre-trained model on the target dataset (FT-A), and 2) only fine-tuning the classifier of the pre-trained model on the target dataset (FT-C) which is equivalent to removing the ICL module from our proposed method. \\
\noindent{\textbf{Metrics.}} We set the metrics for each dataset according to the tasks. For MedFuse-I and MedFuse-P, we use the AUROC and AUPRC as evaluation metrics. For Food-101 and Hateful Memes, we use the accuracy and the AUROC respectively.\\
\noindent{\textbf{Input data processing.}} For MedFuse-I, we use linear embedding to map EHR to token embeddings and the number of embedded EHR tokens is 48. The number of tokens from X-ray image patches is 144. For MedFuse-P, the token numbers are 96 and 96 for EHR and X-rays. The maximum length of text inputs is 512 for the Food-101 task and 128 for Hateful Memes, and the image processing of the input images is the same as \cite{lee2023multimodal}.\\
\noindent{\textbf{Pretrained Multimodal Transformer.}} We use the pre-trained multimodal transformer, ViLT \cite{kim2021vilt}, to extract features following \cite{lee2023multimodal}. For medical data, we use a pre-trained ViLT model and fine-tune all model parameters on one dataset and then use the fine-tuned model on the other dataset. The reason is that there is a huge gap between the pre-training data, i.e., images and texts, and the downstream data, i.e., EHR and X-ray. Thus, fine-tuning all model parameters helps the model to adapt to the medical data. For the Food-101 task and Hateful Memes, we directly use the  ViLT model since there is no such domain gap as in the medical data. For medical datasets, we initialize two kinds of empty tokens and update them in the fine-tuning phase, and then use these tokens on another dataset to represent the missing modality. For Food-101 and HatefulMemes datasets, if the image is missing, we create an image with all pixel values equal to one as dummy input, and if the text is missing, we
use an empty string as dummy input by following \cite{lee2023multimodal}. \\
\noindent{\textbf{ICL module settings.}}\label{ICL module setting.}  We use a 2-layer  transformer and 4 context samples. For computational efficiency, we pool the feature tokens $H$ from ViLT before the input into ICL. The number of pooled tokens is 8. Before the training/testing process, we saved the features inferred by the pretrained multimodal transformer for all full-modality samples in the training set. Then, for each input sample, we first 
obtain its features by the pretrained multi-modal transformer and only use the $cls$ token to retrieve neighbors through the saved features. Finally,  the features of the current sample with its neighbors are input into the ICL module for classification. Note that we only use full-modality training data during the NN search so the computational cost is much less than using all training data.

\begin{table}[t]
\caption{Quantitative results on the MedFuse-I, MedFuse-P, Food-101, and HatefulMemes datasets  under various modality-missing scenarios (here we show the result at $r_{sub}$ = 0.01, see Appendix B for all sample sizes.) The bold number indicates the best performance. F means full-modality, $m_1$ means text/EHR and $m_2$ means image/X-ray. In this scenario, our proposed ICL-CA method outperforms the MAP method by an average of 6.1$\%$ over all datasets.}
\vspace{-0.4cm}
\label{General performance of all methods}
\centering
\resizebox{1\linewidth}{!}{
\begin{tabular}{c|cll|c|ccccc}
\hline
Datasets      & \multicolumn{3}{c|}{Missing state}                                  & Metric       & ICL-CA             & ICL-NTP             & FT-A          & FT-C          & MAP          \\ \hline
MedFuse-I     & \multicolumn{3}{c|}{\tabincell{c}{26\% F, 74\% $m_1$}}              & \tabincell{c}{AUROC\\AUPRC} & \tabincell{c}{\textbf{0.750}\\\textbf{0.308}} & \tabincell{c}{0.737\\0.286} & \tabincell{c}{0.719\\0.257} & \tabincell{c}{0.702\\0.269} & \tabincell{c}{0.691\\0.285} \\ \hline
MedFuse-P     & \multicolumn{3}{c|}{\tabincell{c}{18\% F, 82\% $m_1$}}               & \tabincell{c}{AUROC\\AUPRC} & \tabincell{c}{\textbf{0.556}\\\textbf{0.219}} & \tabincell{c}{0.539\\0.204} & \tabincell{c}{0.504\\0.191} & \tabincell{c}{0.490\\0.189} & \tabincell{c}{0.493\\0.190} \\ \hline
Food-101       & \multicolumn{3}{c|}{\tabincell{c}{30\% F, 70\% $m_2$\\30\% F, 70\% $m_1$\\30\% F, 35\% $m_2$, 35\% $m_1$}} & Accuracy     & \tabincell{c}{0.312\\\textbf{0.342}\\0.281} & \tabincell{c}{\textbf{0.317}\\0.327\\\textbf{0.289}} & \tabincell{c}{0.250\\0.265\\0.228} & \tabincell{c}{0.222\\0.243\\0.196} & \tabincell{c}{0.222\\0.247\\0.186} \\ \hline
HatefulMemes & \multicolumn{3}{c|}{\tabincell{c}{30\% F, 70\% $m_2$\\30\% F, 70\% $m_1$\\30\% F, 35\% $m_2$, 35\% $m_1$}} & AUROC     & \tabincell{c}{\textbf{0.576}\\\textbf{0.577}\\\textbf{0.593}} & \tabincell{c}{0.565\\0.576\\0.583} & \tabincell{c}{0.537\\0.548\\0.539} & \tabincell{c}{0.542\\0.540\\0.532} & \tabincell{c}{0.528\\0.531\\0.529} \\ \hline
Average&\multicolumn{3}{c|}{N.A.}& N.A. & \textbf{0.451}& 0.442& 0.404& 0.393 &0.390 \\ \hline
\end{tabular}
}
\end{table}

\noindent \textbf{Setting of Missing Modalities} For MedFuse-I and MedFuse-P, both of them provide the missing setting: 74$\%$ and 82$\%$ of patients have missing X-rays, respectively. For the other two tasks, we follow the setting in \cite{lee2023multimodal} where three kinds of missing states are designed: 1) 30$\%$ samples have complete modalities and 70$\%$ samples are missing images, 2) 30$\%$ samples have complete modalities and 70$\%$ samples are missing texts, and 3) 30$\%$ samples have complete modalities, 35$\%$ samples are missing images and 35$\%$ samples are missing texts.


\noindent \textbf{Subsampling}.
We subsample a full dataset to simulate a low-data downstream task. For medical data, we subsample the training dataset to 0.01, 0.02, 0.04, 0.1, 0.2, 0.4. For the other two tasks, we subsample the training dataset to 0.01, 0.02, 0.04, 0.1.  We maintain the original missing-modality rate setting during subsampling. We use $r_{sub}$ to refer to the subsampling ratio in the later content.
\vspace{-0.5cm}


\subsection{Main Results}\label{Main Results}
\vspace{-0.1cm}

\begin{figure}[t]

    \centering
    \includegraphics[width=1\linewidth]{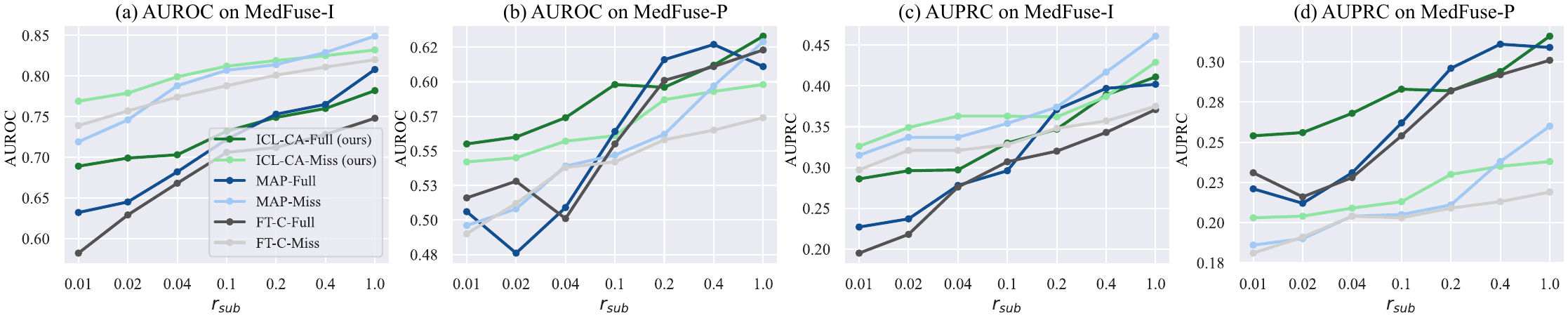}
    \vspace{-0.7cm}
    \caption{The performance of MAP and ICL-CA on MedFuse-I and MedFuse-P when using different training set sizes. Our proposed ICL method is highly competitive under low data cases ($r_{sub}$ from 0.01 to 0.1). Crucially, our approach enhances the performance in both full- and missing-modalities, outperforming the MAP baseline.}
    \label{fig:MedFuse_subsample}
    \vspace{-0.5cm}
\end{figure}

\begin{figure}[t]
    \centering
\includegraphics[width=1\linewidth]{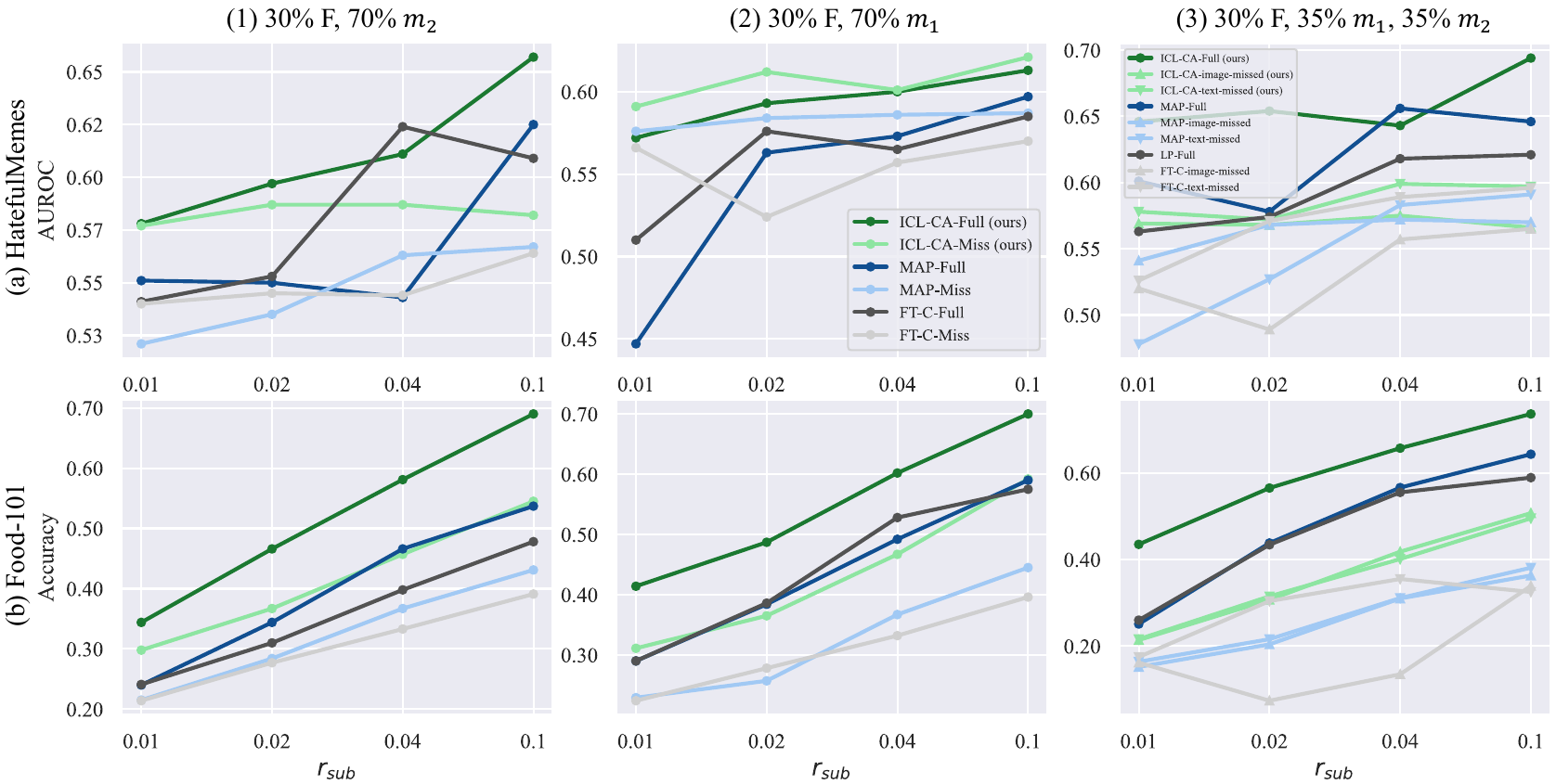}
    \vspace{-0.6cm}
    \caption{The performance of MAP and ICL-CA on HatefulMemes and Food-101 when using different training set sizes. The performance of our ICL-CA is much better than that of MAP in the low-data regime ($r_{sub}$ from 0.01 to 0.1).}
    \vspace{-0.6cm}
    \label{fig:vision_image_subsample}
\end{figure}

Table \ref{General performance of all methods} presents quantitative results across a range of scenarios, where the target dataset's downsampling ratio $r_{sub}$ is 1$\%$. See Appendix B for the performance of all downsampling ratios. From Table \ref{General performance of all methods}, Appendix B, Figures \ref{fig:MedFuse_subsample} and \ref{fig:vision_image_subsample}, we draw the following observations. (1) Across various datasets and scenarios of missing data, a consistent trend emerges: With sufficient target dataset size (notably for $r_{sub} > 0.1$), FT-A exhibits superior performance, attributed to the update of all parameters in the target domain. MAP follows closely, achieving competitive results by updating fewer parameters. In contrast, when the target data is limited, our proposed ICL method, particularly ICL-CA, demonstrates remarkable efficacy (especially for $r_{sub} \leq 0.1$), surpassing most baseline approaches. This trend intensifies as $r_{sub}$ decreases. (2) A notable performance difference is observed between complete and missing modalities across tasks. In simpler binary classification tasks, such as MedFuse-I and HatefulMemes, the performance with full-modality information falls behind that achieved with missing modalities, indicating that complete modality data isn't always necessary for fitting the training data. In contrast, for more complex tasks, such as the multi-label classification in MedFuse-P and the multi-class classification in Food-101, the full-modality data exhibit dominant performance. Our ICL method shows remarkable adaptability under these varying conditions. Furthermore, we observe that ICL  reduces almost all the performance gap between the two modalities, as detailed in Tab.~\ref{tab:performance_gap}.


\begin{table}[t]
\caption{The relative performance gap between missing-modality and full-modality samples on four datasets. The relative performance gap of all settings is shown in Appendix B, where ICL-CA's averaged relative performance gap (22.1\%) is lower than that of MAP (24.1\%). For the exception in Food-101, We hypothesize this is because Food-101 is a high-complex multi-classification task that heavily relies on full-modality data, so using full-modality neighbors in our proposed method enhances the performance of full-modality samples and widens the performance gap.}
\small
\vspace{-0.6cm}
\begin{center}
\begin{tabular}{c|c|c|c|c}
\hline
\multirow{2}{*}{Dataset} & HatefulMemes       & Food-101           & \multirow{2}{*}{MedFuse-I} & \multirow{2}{*}{MedFuse-P} \\ \cline{2-3}
                         & 30\% F, 70\% $m_1$ & 30\% F, 70\% $m_1$ &                            &                            \\ \hline
MAP                &   28.9\%                 &      \textbf{26.6\%}              &            13.8\%                &   2.0\%                         \\ 
\hline
ICL-CA                &   \textbf{3.3\%}                 &   33.1\%                 &      \textbf{11.6\%}                      &    \textbf{1.8\%}                        \\
\hline
\end{tabular}
\end{center}
\vspace{-0.4cm}
\label{tab:performance_gap}
\vspace{-0.1cm}
\end{table}

\subsection{Ablation study}
\noindent{\textbf{ICL by Masked Feature Modeling (ICL-MF).}} 
We explore the efficacy of ICL by employing masked feature modeling with a transformer encoder.  In this approach, we randomly mask a certain number of the input tokens with the mask tensor ($cls_i$ is forced to be masked) and calculate the loss separately. For feature tokens $H$, we calculate the MSE loss between the reconstructed token and the original token. For $cls$ tokens, we train a classifier and compute the loss of the output of the classifier concerning the ground-truth labels. The results of these experiments are in Table \ref{table:mask_encoder}. We compare the performance of ICL-MF against ICL-CA, ICL-NTP and MAP. This comparison is conducted across four distinct dataset settings: MedFuse-I, MedFuse-P, Food-101 (comprising 30$\%$ F and 70$\%$ ${m_1}$ ), and HatefulMemes (with the same missing state as Food-101). Additionally, we evaluate them under two subsampling scenarios, specifically at $r_{sub}$ of 0.01 and 0.1.


Table~\ref{table:mask_encoder}, reveals that ICL-MF either underperforms or marginally surpasses ICL-NTP  and has a clear gap with ICL-CA across all tested settings. It is speculated that this outcome stems from the intrinsic nature of ICL-MF's use of self-attention, which treats each token uniformly. This approach differs from the mechanism employed in cross attention and next-token prediction, which inherently distinguishes between current and similar samples. However, it is noteworthy that ICL-MF demonstrates a significant performance advantage over the MAP approach.

\begin{table}[t]
\small
\centering

\caption{Comparison of ICL-CA, ICL-NTP, ICL-MF and MAP under different datasets. The bold number indicates the best performance.}
\vspace{-0.4cm}
\resizebox{1\linewidth}{!}{
\begin{tabular}{c|c|cccc|cccc}
\hline
\multirow{2}{*}{Datasets} & \multirow{2}{*}{Metric} & \multicolumn{4}{c|}{$r_{sub}= 0.01$} & \multicolumn{4}{c}{$r_{sub}= 0.1$} \\ 
&&ICL-CA    & ICL-NTP & ICL-MF & MAP &ICL-CA & ICL-NTP & ICL-MF & MAP \\ \hline
MedFuse-I & \tabincell{c}{AUROC\\AUPRC} & \tabincell{c}{\textbf{0.750}\\\textbf{0.308}} & \tabincell{c}{0.737\\0.286} & \tabincell{c}{0.733\\0.293}  & \tabincell{c}{0.691\\0.285}& \tabincell{c}{\textbf{0.793}\\0.352} & \tabincell{c}{0.789\\\textbf{0.355}} & \tabincell{c}{0.791\\0.349} & \tabincell{c}{0.788\\0.338} \\ \hline
MedFuse-P & \tabincell{c}{AUROC\\AUPRC} &\tabincell{c}{\textbf{0.556}\\\textbf{0.219}} &\tabincell{c}{0.539\\0.204}  & \tabincell{c}{0.526\\0.199} & \tabincell{c}{0.493\\0.190} & \tabincell{c}{\textbf{0.578}\\\textbf{0.234}}& \tabincell{c}{0.565\\0.224} & \tabincell{c}{0.569\\0.228} & \tabincell{c}{0.561\\0.224}\\ \hline
\tabincell{c}{Food-101\\30$\%$ F, 70$\%$ $m1$} & Accuracy  &\tabincell{c}{\textbf{0.342}}   & \tabincell{c}{0.327} & \tabincell{c}{0.326} & \tabincell{c}{0.247} & \tabincell{c}{\textbf{0.625}} & \tabincell{c}{0.619} & \tabincell{c}{0.564} & \tabincell{c}{0.489}\\ \hline
\tabincell{c}{HatefulMemes\\30$\%$ F, 70$\%$ $m1$}  & AUROC &\tabincell{c}{\textbf{0.577}} & \tabincell{c}{0.576} & \tabincell{c}{0.561}  & \tabincell{c}{0.531} & \tabincell{c}{\textbf{0.617}} & \tabincell{c}{0.612} & \tabincell{c}{0.608}  & \tabincell{c}{0.586}\\ \hline
\end{tabular}
}
\vspace{-0.3cm}
\label{table:mask_encoder}
\end{table}

\begin{figure} [t]
        \centering
	\includegraphics[width=1.0\linewidth]{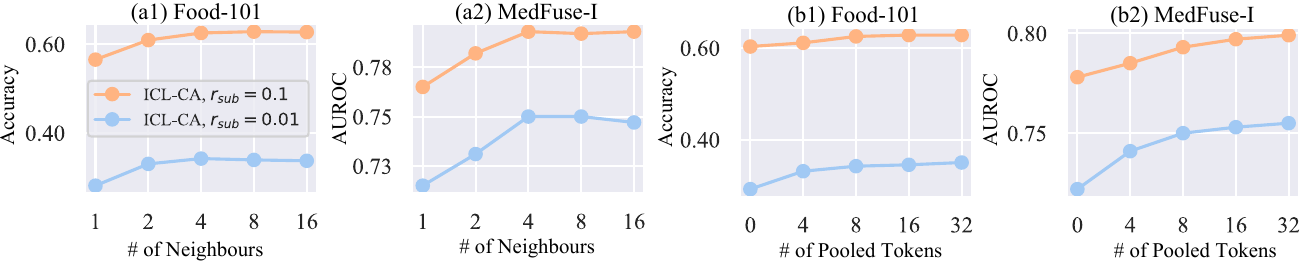}
 \vspace{-0.7cm}
	\caption{Comparison of the effect of the number of neighbors and the pooled feature length in the ICL-CA model. (a) Comparison of the effect of the number of neighbors. (b) Comparison of the effect of pooled feature length. We suggest setting the number of neighbors to 4 and the pooled feature length to 8. } 
	\label{fig:num_of_nn}
  \vspace{-0.2cm}
\end{figure}

\noindent\textbf{The Impact of the Number of Neighbors $Q$.} 
We examine the influence of the number of neighbors on our ICL-CA, as depicted in Fig.~\ref{fig:num_of_nn}a. In this analysis, we vary the number of neighbors (1, 2, 4, 8, 16) and observed their effects on two datasets, MedFuse-I and Food-101, under $r_{sub}$ of 0.1 and 0.01. Our findings indicate a marked performance improvement when the number of neighbors is increased from 1 to 4 in all experiments. However, further increases in the number of neighbors do not sustain this upward trend in performance. Therefore, we use $Q$=4 in our experiment to strike a balance between computational efficiency and efficacy. 



\noindent{\textbf{The Effect of Pooled Feature Length $T$}.} 
We assess the impact of varying pooled feature lengths (the number of pooled feature tokens in each sample) on our ICL-CA, as illustrated in Fig. \ref{fig:num_of_nn}b.    Pooled features of greater length can provide more comprehensive feature information but concurrently increase computational demands.  We test pooled feature lengths of 0, 4, 8, 16, and 32 under $r_{sub}$ of 0.01 and 0.1 in the MedFuse-I and Food-101 datasets. A pooled feature length of 0 implies reliance solely on the $cls$ token from all samples for ICL. A substantial increase in performance is observed when the pooled feature length is increased from 0 to 8. When the pooled feature length exceeds 8, the gain in performance becomes negligible. Thus, we set the pooled feature length to 8 in this paper.

\noindent{\textbf{Groups for retrieving neighboring samples.}  We compare different groups of samples in training datasets for retrieving neighboring samples, i,e. all the samples, full-modality samples and missing-modality samples. We select the ICL-CA method under $r_{sub}=0.01$ for this experiment, as shown in Table \ref{tab: retrieving groups exp}. It shows that employing samples with full modality as the retrieval group yields superior results. This observation indicates that our proposed ICL method can effectively utilize the context provided by the full-modality samples. In addition, using full-modality samples as neighbors enhances computational efficiency due to the reduced  sample size of neighbors.

\begin{table}[t]
\caption{Performance of  ICL-CA by different groups for retrieving neighboring samples under $r_{sub}=0.01$. NN-all, NN-full and NN-miss refer to using all training data, full-modality data and missing-modality ones respectively, in the retrieval process. }
\vspace{-0.3cm}
\label{tab: retrieving groups exp}
\centering
\begin{tabular}{c|cll|c|ccc}
\hline
Datasets      & \multicolumn{3}{c|}{Missing state}                                  & Metric       & NN-all     &    NN-full  &NN-miss            \\ \hline
MedFuse-I     & \multicolumn{3}{c|}{\tabincell{c}{26\% F, 74\% $m_1$}}              & \tabincell{c}{AUROC\\AUPRC} & \tabincell{c}{0.732\\0.269}& \tabincell{c}{\textbf{0.750}\\\textbf{0.308}}& \tabincell{c}{0.721\\0.248} \\ \hline
MedFuse-P     & \multicolumn{3}{c|}{\tabincell{c}{18\% F, 82\% $m_1$}}               & \tabincell{c}{AUROC\\AUPRC} & \tabincell{c}{0.533\\0.191}& \tabincell{c}{\textbf{0.556}\\\textbf{0.219}}& \tabincell{c}{0.520\\0.184}\\ \hline
Food-101       & \multicolumn{3}{c|}{\tabincell{c}{30\% F, 70\% $m_2$\\30\% F, 70\% $m_1$\\30\% F, 35\% $m_2$, 35\% $m_1$}} & Accuracy     & \tabincell{c}{0.294\\\textbf{0.346}\\0.266}& \tabincell{c}{\textbf{0.312}\\0.342\\\textbf{0.281}} & \tabincell{c}{0.265\\0.311\\0.247} \\ \hline
HatefulMemes & \multicolumn{3}{c|}{\tabincell{c}{30\% F, 70\% $m_2$\\30\% F, 70\% $m_1$\\30\% F, 35\% $m_2$, 35\% $m_1$}} & AUROC     & \tabincell{c}{0.549\\0.569\\0.575}  &\tabincell{c}{\textbf{0.576}\\\textbf{0.577}\\\textbf{0.593}} & \tabincell{c}{0.523\\0.553\\0.566}\\ \hline
Average&\multicolumn{3}{c|}{N.A.}& N.A. & 0.432 & \textbf{0.451} &0.414\\ \hline
\end{tabular}
\vspace{-0.5cm}
\end{table}

\noindent\textbf{Inference Time and parameters number. } One major concern for the retrieval-based approach is the inference latency. We test the inference time of ICL-CA and MAP on  MedFuse-I. We set the batch size to 1 and record the inference time for 100 batches. The average inference time of ICL-CA is 34.41ms with a std of 4.52ms. In contrast, MAP has a mean inference time of 40.59ms and a std of 5.40ms. The difference in inference time is because MAP has a larger number of tokens (missing-aware prompts) in the transformer. We also calculate the number of trainable parameters in Appendix B.
\vspace{-0.3cm}

\section{Conclusion}
\label{sec:conclusion}
\vspace{-0.3cm}
This paper investigates a pivotal challenge in multimodal learning: missing modalities in the low-data regime. Our analysis examines the learning process of both full and missing modalities across tasks of various complexity. Stemming from our findings, we introduce a semi-parametric, retrieval-augmented in-context learning framework to address the challenges. This approach is designed to condition each sample with neighboring full-modality data. The effectiveness of our method is corroborated across diverse datasets, including medical and vision-language prediction tasks. Remarkably, our approach achieves an average performance boost of 6.1$\%$ over the baseline in the low-data regime. Furthermore, it effectively narrows the performance disparity caused by modality absence. 
Our future work will focus on efficient and accurate retrieval methods. We will also extend the method to tasks with more modalities.

\clearpage  

%
%
\bibliographystyle{splncs04}
\bibliography{main}
\clearpage
\setcounter{page}{1}
\setcounter{section}{0}
{\centering
        \Large
        \textbf{
        Appendix} \\
}
\renewcommand{\thesection}
{\Alph{section}} 
\section{Experimental Settings}
\noindent{\textbf{Details of datasets}}
The details of each dataset are as follows:\\
$\bullet$ \textit{MedFuse-In-hospital mortality (MedFuse-I) \cite{hayat2022medfuse}.} This dataset contains EHR and X-ray data for each patient. The target of this binary classification task is to predict in-hospital mortality after the first 48 hours spent in the ICU. The EHR is time-series data with 17 clinical variables, among which five are categorical and 12 are continuous.  Each EHR is paired with the last chest X-ray image collected during the ICU stay. The numbers of the samples in the training/val/testing dataset are 18845, 2138 and 5243.\\
$\bullet$ \textit{MedFuse-Phenotype (MedFuse-P)  \cite{hayat2022medfuse}} This dataset has the same data types as in MedFuse-I. The difference is that this dataset has a larger sample size and the task is multi-label classification to predict whether a set of 25 chronic, mixed, and acute care conditions are assigned to a patient in a given ICU stay. The numbers of the samples in the training/val/testing dataset are 42628, 4802 and 11914.\\
$\bullet$ \textit{UPMC Food-101 \cite{bossard14}}. This dataset contains the noisy text-image paired data for 101 kinds of food. The target is to predict the type of food, which is a multi-classification task. The numbers of the samples in the training/val/testing dataset are 61127, 6588 and 25250.\\
$\bullet$ \textit{Hateful Memes \cite{kiela2020hateful}}. This is a binary classification task. The dataset represents a challenging blend of visual and textual content, specifically designed to tackle the detection of harmful content online. The dataset comprises meme images that are often used in social media contexts, containing layers of nuances in meaning that combine text and imagery. The numbers of the samples in the training/val/testing dataset are 8500, 500 and 1000.

\section{More Experimental Results}
\noindent{\textbf{The performance of our method and baselines on \emph{all test samples}}}. The main paper presents the performance on full-modality and missing-modality test samples \emph{separately} in the figures. Here we give the performance of our method and baselines on \emph{all test samples} as in \cite{lee2023multimodal}. Table \ref{performance of all methods big than 0.1} presents quantitative results of all test samples across all datasets, methods, and missing states under $r_{sub}\geq 0.1$.  Table \ref{performance of all methods small than 0.1} presents the results under $r_{sub}<0.1$.

\noindent{\textbf{The relative performance gap between missing- and full-modality on all datasets}}. The main paper presents the relative performance gap, i.e., $(\text{Metric}_{\text{high}}-\text{Metric}_{\text{low}})/\text{Metric}_{\text{low}}$), between missing-modality and full-modality data on four missing states under $r_{sub} = 0.01$. Here we show the details of the performance of missing-modality and full-modality data on all missing states and all datasets in Table \ref{details of gap}.

\noindent{\textbf{The number of trainable parameters of our methods and baselines}}. We also calculate the number of trainable parameters for all methods, as shown in Table \ref{tab: comparison of num of parameters}. Although our method has more parameters than MAP, it is less overfitting to the training data in the low-data regime compared with MAP.

\begin{table}[htbp]
\caption{Quantitative results of the whole test set on the Medfuse-I, Medfuse-P, Food101, and HatefulMemes datasets with different missing rates under various modality-missing scenarios under $r_{sub}\geq 0.1$. Bold number indicates the best performance. With sufficient target dataset size (notably for $r_{sub} > 0.1$), FT-A exhibits superior performance, attributed to the update of all parameters in the target domain. MAP follows closely, achieving competitive results by updating fewer parameters. FT-C, on the other hand, performs the worst at all moments, due to the limited number of updated parameters.}\label{performance of all methods big than 0.1}
\centering
\small
\resizebox{\linewidth}{!}{
\begin{tabular}{c|c|c|c|ccccc}
\hline
$r_{sub}$& Datasets      & Missing state &Metric & ICL-CA             & ICL-NTP             & FT-A          & FT-C          & MAP \\ \hline
\multirow{8}{*}{0.1}& \tabincell{c}{Medfuse-I}     & \tabincell{c}{26\% F, 74\% $m1$}      
& \tabincell{c}{AUROC\\AUPRC}& \tabincell{c}{\textbf{0.793}\\0.352} & \tabincell{c}{0.789\\0.355} & \tabincell{c}{0.790\\\textbf{0.356}} & \tabincell{c}{0.771\\0.321} & \tabincell{c}{0.788\\0.338} \\ \cline{2-9}
&\tabincell{c}{Medfuse-P}    & \tabincell{c}{18\% F, 82\% $m1$}           & \tabincell{c}{AUROC\\AUPRC}   & \tabincell{c}{0.578\\0.234} & \tabincell{c}{0.565\\0.224} & \tabincell{c}{\textbf{0.592}\\\textbf{0.246}} & \tabincell{c}{0.556\\0.219} & \tabincell{c}{0.561\\0.224} \\ \cline{2-9}
&\tabincell{c}{Food101 }      & \tabincell{c}{30\% F, 70\% $m2$\\30\% F, 70\% $m1$\\30\% F, 35\% $m1$, 35\% $m2$}  & \tabincell{c}{Accuracy}   & \tabincell{c}{\textbf{0.595}\\\textbf{0.625}\\\textbf{0.571}} & \tabincell{c}{0.576\\0.619\\0.566} & \tabincell{c}{0.562\\0.603\\0.535} & \tabincell{c}{0.417\\0.450\\0.409} & \tabincell{c}{0.463\\0.489\\0.453} \\ \cline{2-9}
&\tabincell{c}{Hateful\\Memes} & \tabincell{c}{30\% F, 70\% $m2$\\30\% F, 70\% $m1$\\30\% F, 35\% $m1$, 35\% $m2$} &  \tabincell{c}{AUROC}     & \tabincell{c}{\textbf{0.607}\\\textbf{0.617}\\\textbf{0.618}} & \tabincell{c}{0.598\\0.612\\\textbf{0.618}} & \tabincell{c}{0.601\\0.609\\0.614} & \tabincell{c}{0.577\\0.575\\0.579} & \tabincell{c}{0.585\\0.586\\0.599} \\ \hline
\multirow{4}{*}{0.2}& \tabincell{c}{Medfuse-I}     & \tabincell{c}{26\% F, 74\% $m1$}      
& \tabincell{c}{AUROC\\AUPRC}& \tabincell{c}{0.802\\0.353} & \tabincell{c}{0.792\\0.370} & \tabincell{c}{\textbf{0.832}\\\textbf{0.411}} & \tabincell{c}{0.782\\0.339} & \tabincell{c}{0.801\\0.370} \\ \cline{2-9}
&\tabincell{c}{Medfuse-P}    & \tabincell{c}{18\% F, 82\% $m1$}           & \tabincell{c}{AUROC\\AUPRC}   & \tabincell{c}{0.590\\0.243} & \tabincell{c}{0.580\\0.231} & \tabincell{c}{\textbf{0.651}\\\textbf{0.290}} & \tabincell{c}{0.576\\0.232} & \tabincell{c}{0.581\\0.237} \\ \hline
\multirow{4}{*}{0.4}& \tabincell{c}{Medfuse-I}     & \tabincell{c}{26\% F, 74\% $m1$}      
& \tabincell{c}{AUROC\\AUPRC}& \tabincell{c}{0.810\\0.388} & \tabincell{c}{0.806\\0.399} & \tabincell{c}{\textbf{0.840}\\\textbf{0.451}} & \tabincell{c}{0.793\\0.351} & \tabincell{c}{0.815\\0.410} \\ \cline{2-9}
&\tabincell{c}{Medfuse-P}    & \tabincell{c}{18\% F, 82\% $m1$}           & \tabincell{c}{AUROC\\AUPRC}   & \tabincell{c}{0.602\\0.251} & \tabincell{c}{0.593\\0.242} & \tabincell{c}{\textbf{0.688}\\\textbf{0.325}} & \tabincell{c}{0.583\\0.237} & \tabincell{c}{0.609\\0.259} \\ \hline
\multirow{4}{*}{1.0}& \tabincell{c}{Medfuse-I}     & \tabincell{c}{26\% F, 74\% $m1$}      
& \tabincell{c}{AUROC\\AUPRC}& \tabincell{c}{0.820\\0.420} & \tabincell{c}{0.819\\0.431} & \tabincell{c}{\textbf{0.850}\\\textbf{0.490}} & \tabincell{c}{0.804\\0.372} & \tabincell{c}{0.838\\0.444} \\ \cline{2-9}
&\tabincell{c}{Medfuse-P}    & \tabincell{c}{18\% F, 82\% $m1$}           & \tabincell{c}{AUROC\\AUPRC}   & \tabincell{c}{0.611\\0.261} & \tabincell{c}{0.596\\0.244} & \tabincell{c}{\textbf{0.704}\\\textbf{0.347}} & \tabincell{c}{0.591\\0.243} & \tabincell{c}{0.630\\0.273} \\ \hline
\end{tabular}
}
\end{table}

\begin{table}[htbp]
\caption{Quantitative results of the whole test set on the Medfuse-I, Medfuse-P, Food101, and HatefulMemes datasets with different missing rates under various modality-missing scenarios. Bold number indicates the best performance. When the target data is limited, our proposed ICL method, particularly ICL-CA, demonstrates remarkable efficacy (especially for $r_{sub} < 0.1$), surpassing most baseline approaches. This trend intensifies as $r_{sub}$ decreases. }\label{performance of all methods small than 0.1}
\centering
\small
\resizebox{\linewidth}{!}{
\begin{tabular}{c|c|c|c|ccccc}
\hline
$r_{sub}$& Datasets      & Missing state &Metric & ICL-CA             & ICL-NTP             & FT-A          & FT-C          & MAP \\ \hline
\multirow{9}{*}{0.01}& \tabincell{c}{Medfuse-I}     & \tabincell{c}{26\% F, 74\% $m1$}      
& \tabincell{c}{AUROC\\AUPRC}& \tabincell{c}{\textbf{0.750}\\\textbf{0.308}} & \tabincell{c}{0.737\\0.286} & \tabincell{c}{0.719\\0.257} & \tabincell{c}{0.702\\0.269} & \tabincell{c}{0.691\\0.285} \\ \cline{2-9}
&\tabincell{c}{Medfuse-P}    & \tabincell{c}{18\% F, 82\% $m1$}           & \tabincell{c}{AUROC\\AUPRC}   & \tabincell{c}{\textbf{0.556}\\\textbf{0.219}} & \tabincell{c}{0.539\\0.204} & \tabincell{c}{0.504\\0.191} & \tabincell{c}{0.490\\0.189} & \tabincell{c}{0.493\\0.190} \\ \cline{2-9}
&\tabincell{c}{Food101 }      & \tabincell{c}{30\% F, 70\% $m2$\\30\% F, 70\% $m1$\\30\% F, 35\% $m1$, 35\% $m2$}  & \tabincell{c}{Accuracy}   & \tabincell{c}{0.312\\\textbf{0.342}\\0.281} & \tabincell{c}{\textbf{0.317}\\0.327\\\textbf{0.289}} & \tabincell{c}{0.250\\0.265\\0.228} & \tabincell{c}{0.222\\0.243\\0.196} & \tabincell{c}{0.222\\0.247\\0.186} \\ \cline{2-9}
&\tabincell{c}{Hateful\\Memes} & \tabincell{c}{30\% F, 70\% $m2$\\30\% F, 70\% $m1$\\30\% F, 35\% $m1$, 35\% $m2$} &  \tabincell{c}{AUROC}     & \tabincell{c}{\textbf{0.576}\\\textbf{0.577}\\\textbf{0.593}} & \tabincell{c}{0.565\\0.576\\0.583} & \tabincell{c}{0.537\\0.548\\0.539} & \tabincell{c}{0.542\\0.540\\0.532} & \tabincell{c}{0.528\\0.531\\0.529} \\ \hline
\multirow{8}{*}{0.02}& 
\tabincell{c}{Medfuse-I}     & \tabincell{c}{26\% F, 74\% $m1$}   & \tabincell{c}{AUROC\\AUPRC}& \tabincell{c}{0.761\\\textbf{0.328}} & \tabincell{c}{\textbf{0.764}\\\textbf{0.328}} & \tabincell{c}{0.754\\0.299} & \tabincell{c}{0.728\\0.293} & \tabincell{c}{0.722\\0.308} \\ \cline{2-9}
&\tabincell{c}{Medfuse-P}    & \tabincell{c}{18\% F, 82\% $m1$}           & \tabincell{c}{AUROC\\AUPRC}   & \tabincell{c}{\textbf{0.559}\\\textbf{0.221}} & \tabincell{c}{0.552\\0.212} & \tabincell{c}{0.530\\0.207} & \tabincell{c}{0.524\\0.198} & \tabincell{c}{0.522\\0.198} \\ \cline{2-9}
&\tabincell{c}{Food101 }      & \tabincell{c}{30\% F, 70\% $m2$\\30\% F, 70\% $m1$\\30\% F, 35\% $m1$, 35\% $m2$}  & \tabincell{c}{Accuracy}   & \tabincell{c}{\textbf{0.397}\\\textbf{0.402}\\\textbf{0.387}} & \tabincell{c}{0.373\\0.389\\0.355} & \tabincell{c}{0.352\\0.338\\0.315} & \tabincell{c}{0.287\\0.310\\0.262} & \tabincell{c}{0.302\\0.295\\0.278} \\ \cline{2-9}
&\tabincell{c}{Hateful\\Memes} & \tabincell{c}{30\% F, 70\% $m2$\\30\% F, 70\% $m1$\\30\% F, 35\% $m1$, 35\% $m2$} &  \tabincell{c}{AUROC}     & \tabincell{c}{\textbf{0.590}\\\textbf{0.593}\\0.602} & \tabincell{c}{0.581\\0.587\\\textbf{0.603}} & \tabincell{c}{0.550\\0.570\\0.564} & \tabincell{c}{0.545\\0.556\\0.549} & \tabincell{c}{0.545\\0.557\\0.548} \\ \hline
\multirow{8}{*}{0.04}& 
\tabincell{c}{Medfuse-I}     & \tabincell{c}{26\% F, 74\% $m1$}      
& \tabincell{c}{AUROC\\AUPRC}& \tabincell{c}{0.778\\\textbf{0.344}} & \tabincell{c}{0.777\\0.336} & \tabincell{c}{\textbf{0.787}\\0.328} & \tabincell{c}{0.752\\0.308} & \tabincell{c}{0.767\\0.319} \\ \cline{2-9}
&\tabincell{c}{Medfuse-P}    & \tabincell{c}{18\% F, 82\% $m1$}           & \tabincell{c}{AUROC\\AUPRC}   & \tabincell{c}{\textbf{0.569}\\\textbf{0.228}} & \tabincell{c}{0.557\\0.219} & \tabincell{c}{0.561\\0.213} & \tabincell{c}{0.542\\0.209} & \tabincell{c}{0.545\\0.211} \\ \cline{2-9}
&\tabincell{c}{Food101 }      & \tabincell{c}{30\% F, 70\% $m2$\\30\% F, 70\% $m1$\\30\% F, 35\% $m1$, 35\% $m2$}  & \tabincell{c}{Accuracy}   & \tabincell{c}{\textbf{0.494}\\\textbf{0.508}\\\textbf{0.484}} & \tabincell{c}{0.464\\0.489\\0.460} & \tabincell{c}{0.448\\0.458\\0.422} & \tabincell{c}{0.352\\0.391\\0.338} & \tabincell{c}{0.397\\0.405\\0.387} \\ \cline{2-9}
&\tabincell{c}{Hateful\\Memes} & \tabincell{c}{30\% F, 70\% $m2$\\30\% F, 70\% $m1$\\30\% F, 35\% $m1$, 35\% $m2$} &  \tabincell{c}{AUROC}     & \tabincell{c}{\textbf{0.595}\\\textbf{0.600}\\0.601} & \tabincell{c}{0.591\\0.585\\\textbf{0.607}} & \tabincell{c}{0.583\\0.577\\0.579} & \tabincell{c}{0.561\\0.559\\0.562} & \tabincell{c}{0.558\\0.567\\0.574} \\ \hline
\end{tabular}
}
\end{table}

\begin{table}[htbp]
\caption{The relative performance gap between missing-modality and full-modality data on four datasets and all missing states under $r_{sub}=0.01$. Bold number indicates the best performance. In most of the settings, our proposed ICL-CA shows smaller relative gap compared to the baseline MAP. ICL-CA’s averaged relative performance gap (22.1$\%$) is lower than that of
MAP (24.1$\%$).  }\label{details of gap}
\centering
\small
\resizebox{\linewidth}{!}{
\begin{tabular}{c|c|c|ccc|ccc}
\hline
&&& \multicolumn{3}{c|}{ICL-CA} & \multicolumn{3}{c}{MAP} \\ \hline  
Dataset&Missing state&Metric & \tabincell{c}{missing \\modality} & \tabincell{c}{full\\ modality} & \tabincell{c}{relative\\ gap ($\%$)} & \tabincell{c}{missing \\modality} & \tabincell{c}{full\\ modality} & \tabincell{c}{relative\\ gap ($\%$)}\\ \hline

Medfuse-I&\tabincell{c}{26\% F, 74\% $m1$}      
& \tabincell{c}{AUROC\\AUPRC}&\tabincell{c}{0.769\\0.326} & \tabincell{c}{0.689\\0.286}& \tabincell{c}{\textbf{11.6}\\\textbf{14.0}} & \tabincell{c}{0.719\\0.337}& \tabincell{c}{0.632\\0.237}&\tabincell{c}{13.8\\42.2} \\ \hline

\tabincell{c}{Medfuse-P}    & \tabincell{c}{18\% F, 82\% $m1$}           & \tabincell{c}{AUROC\\AUPRC} &\tabincell{c}{0.545\\0.203} & \tabincell{c}{0.555\\0.254}&\tabincell{c}{\textbf{1.8}\\25.1}&\tabincell{c}{0.496\\0.186}&\tabincell{c}{0.506\\0.221}&\tabincell{c}{2.0\\\textbf{18.8}} \\ \hline

\tabincell{c}{Food101 }      & \tabincell{c}{30\% F, 70\% $m2$\\30\% F, 70\% $m1$\\30\% F, 35\% $m1$, 35\% $m2$}  & \tabincell{c}{Accuracy} &\tabincell{c}{0.298\\0.311\\0.214} &\tabincell{c}{0.344\\0.414\\0.435}&\tabincell{c}{15.4\\33.1\\103.3}&\tabincell{c}{0.215\\0.229\\0.152}&\tabincell{c}{0.240\\0.290\\0.251}&\tabincell{c}{\textbf{11.6}\\\textbf{26.6}\\\textbf{65.1}} \\ \hline

\tabincell{c}{Hateful\\Memes }      & \tabincell{c}{30\% F, 70\% $m2$\\30\% F, 70\% $m1$\\30\% F, 35\% $m1$, 35\% $m2$}  & \tabincell{c}{AUROC} &\tabincell{c}{0.577\\0.591\\0.569} &\tabincell{c}{0.578\\0.572\\0.646}&\tabincell{c}{\textbf{0.2}\\\textbf{3.3}\\\textbf{13.5}}&\tabincell{c}{0.521\\0.576\\0.478}&\tabincell{c}{0.551\\0.447\\0.601}&\tabincell{c}{5.8\\28.9\\25.7} \\ \hline

Average&&&&&\textbf{22.1}& &&24.1\\ \hline

\end{tabular}
}
\end{table}
\begin{table}[t]
\caption{Number of trainable parameters for different models.}
\label{tab: comparison of num of parameters}
    \centering
    \begin{tabular}{ccccc}
    \hline
        Model& MAP & ICL-CA &FT-A &FT-C \\
        num/million&2.8 & 5.1 &113 &0.6 \\
        \hline
    \end{tabular}
\end{table}

\end{document}